\documentclass{Interspeech}

\interspeechcameraready 

\usepackage{threeparttable} 

\title{Benchmarking Training Paradigms, Dataset Composition, and Model Scaling for Child ASR in ESPnet}

\author[affiliation={1}]{Anyu}{Ying}
\author[affiliation={2}]{Natarajan Balaji}{Shankar}
\author[affiliation={1}]{Chyi-Jiunn}{Lin}
\author[affiliation={2}]{Mohan}{Shi}
\author[affiliation={3}]{Pu}{Wang}
\author[affiliation={1}]{Hye-jin}{Shim}
\author[affiliation={1}]{Siddhant}{Arora}
\author[affiliation={3}]{Hugo}{Van hamme}
\author[affiliation={2}]{Abeer}{Alwan}
\author[affiliation={1}]{Shinji}{Watanabe}

\affiliation{}{Carnegie Mellon University}{USA}
\affiliation{}{University of California Los Angeles}{USA}
\affiliation{}{KU Leuven}{Belgium}
\email{anyuy@alumni.cmu.edu, balaji1312@ucla.edu, chyjiul@andrew.cmu.edu,\\ shimohan@g.ucla.edu, pu.wang@esat.kuleuven.be, shimhz6.6@gmail.com,\\ siddhana@andrew.cmu.edu, hugo.vanhamme@esat.kuleuven.be, alwan@ee.ucla.edu, shinjiw@ieee.org}

\usepackage{comment}
\newcounter{rownumber} 
\newcommand{\rownumber}{\stepcounter{rownumber}\arabic{rownumber}} 

\begin{document}

\maketitle

\begin{abstract}
    
    Despite advancements in ASR, child speech recognition remains challenging due to acoustic variability and limited annotated data. While fine-tuning adult ASR models on child speech is common, comparisons with flat-start training remain underexplored. We compare flat-start training across multiple datasets, SSL representations (WavLM, XEUS), and decoder architectures. Our results show that SSL representations are biased toward adult speech, with flat-start training on child speech mitigating these biases. We also analyze model scaling, finding consistent improvements up to 1B parameters, beyond which performance plateaus. Additionally, age-related ASR and speaker verification analysis highlights the limitations of proprietary models like Whisper, emphasizing the need for open-data models for reliable child speech research. All investigations are conducted using ESPnet, and our publicly available benchmark provides insights into training strategies for robust child speech processing.
    
\end{abstract}

\section{Introduction}

Large-scale speech foundation models have significantly advanced automatic speech recognition (ASR), enabling systems to generalize across diverse speech domains with minimal adaptation \cite{chen2022wavlm, hsu2021hubert, baevski2020wav2vec, Rad23whisper, peng2024owsm, puvvada2024less}. While these advancements have yielded impressive results for adult ASR, child ASR remains an ongoing challenge due to the acoustic and linguistic differences between child and adult speech \cite{lee1999acoustics, yeung2018difficulties, dutta2022challenges}, as well as the limited availability of large, high-quality annotated child speech datasets. Traditional improvements in child ASR rely on data augmentation techniques like vocal tract length perturbation \cite{jaitly2013vocal}, pitch perturbation \cite{patel2011prosodic}, and synthetic data \cite{shahnawazuddin20_interspeech, rolland2024improved}, as well as fine-tuning models pre-trained on adult speech \cite{shivakumar2020transfer, rolland2024improved, fan2024benchmarking, attia2024kid}.

Despite the prevalence of fine-tuning, training child ASR models from scratch remains underexplored. Flat-start training learns child-specific patterns without adult speech biases, while fine-tuning may lead to suboptimal adaptation. By systematically evaluating flat-start models across datasets and architectures, we assess their feasibility compared to fine-tuning. Front-end feature extraction is crucial for ASR robustness. While most child ASR models use raw waveforms or Mel spectrograms, self supervised learning (SSL) approaches may improve performance \cite{shi24g_interspeech, arora-etal-2024-evaluation, zaiem2025speech, li2024analysis, fan2022towards}, especially given child speech's prosodic and phonetic variability. Finally, we examine the impact of continuous vs. discrete speech representations \cite{YangSDM0P024, ChangYFM023} on model performance, evaluating whether discretization benefits child speech modeling.

Another factor influencing ASR performance is dataset composition.  Many child ASR models are trained on homogeneous corpora with homogeneous speaking styles or narrow age distributions, limiting cross-domain performance \cite{attia2024kid, jain2023wav2vec2}. We examine whether multi-corpus training improves generalization, exploring trade-offs between single- and multi-corpus approaches and their impact on dataset bias mitigation and robustness. We conduct an analysis of the effect of age, investigating how model performance varies across different age groups. To better understand age-related variability in speech processing, we also conduct a study on speaker verification, evaluating how speaker embeddings perform across different age groups. In addition to ASR, speaker verification performance is also affected by age-related variability, with younger children's speech exhibiting higher error rates \cite{shahnawazuddin2021children, singh2024childaugment, aziz2023effective}. To understand these effects, we evaluate speaker embeddings across age groups, analyzing speaker verification performance in child speech.

To systematically analyze the impact of model scaling and training data scaling on child ASR performance, we vary both model size and the amount of training data. For model scaling, we evaluate Whisper models \cite{Rad23whisper} ranging from 39 million to 1.55 billion parameters and Open Whisper-style Large-scale neural model Suite (OWLS) models \cite{chen2025owlsscalinglawsmultilingual} from 0.25 to 9 billion parameters in a zero-shot setting. 

To facilitate reproducibility, all experiments are conducted within an ESPnet-based framework \cite{watanabe18_interspeech}, evaluating a variety of open-source models, including WavLM \cite{chen2022wavlm}, OWSM \cite{peng2023reproducing}, and XEUS \cite{chen2024towards}. Our primary focus is on open-data foundation models to mitigate biases and risks of data leakage, though we include Whisper for comparative analysis.

We sum up our contributions as follows:

\begin{itemize}
\item 
We offer the first systematic comparison of flat-start training versus fine-tuning pre-trained ASR models across multiple child speech datasets. Additionally, we investigate how front-end representations, model architectures, and representation types (continuous vs. discrete) impact performance, providing guidance on optimal configurations for robustness.
\item 
We evaluate the impact of single-corpus vs. multi-corpus training on ASR performance, highlighting trade-offs between domain-specific adaptation and cross-domain generalization. Additionally, we analyze performance of ASR and speaker verification systems across different age groups, examining how model architectures handle various developmental stages of child speech.
\item 
We analyze the impact of model scaling on child ASR by evaluating Whisper (39M–1.55B) and OWLS (0.25–9B) models. 
\item 
Experiments are conducted in a unified ESPnet-based framework. We provide complete recipes for each corpus, detailing data preparation, model training, and evaluation. Our database-specific data processing ensures fair comparisons, with all treatments carefully documented. This thorough documentation ensures reproducibility and serves as a strong benchmark for future child speech processing research\footnote{\url{https://github.com/espnet/espnet/tree/master/egs2/myst_ogi_cmu_kids}}.

\end{itemize}
\vspace{-5pt}
\section{Methodology}
We conduct controlled experiments to investigate key factors affecting child ASR performance: training strategies (flat-start vs. fine-tuning), front-end representations, representation types (continuous vs. discrete), dataset composition (single- vs. multi-corpus), model scaling, and age-based performance of ASR and speaker verification systems.

\vspace{-5pt}
\subsection{Flat-start training vs. fine-tuning}
\vspace{-5pt}
For flat-start-based models, we train ASR models with three different decoder strategies using a 12-layer E-Branchformer \cite{kim2023branchformer} as the encoder:

\begin{itemize}
\item \textbf{Attention-based encoder-decoder (AED)} \cite{karita2019comparative}: 6-layer Transformer decoder with 4 attention heads and 2048 linear units, trained with hybrid CTC (weight 0.3).

\item \textbf{RNN transducer (RNNT)} \cite{Graves2012SequenceTW}: Single-layer LSTM prediction network (256 units) with a 320-dimensional joint network, trained with hybrid CTC (weight 0.3).

\item \textbf{Connectionist temporal classification (CTC)} \cite{graves2006ctc}: CTC 
  output layer applied to the encoder. We evaluate both 
  standalone CTC and CTC enhanced with a 2-layer LSTM 
  external language model (650 hidden units per layer) 
  trained on the same data as the ASR model.
\end{itemize}

For fine-tuning-based models, we adapt:
\begin{itemize}
    \item \textbf{OWSM v3.1} \cite{peng2024owsm}: An AED model trained on 180K hours of public data, featuring an 18-layer E-Branchformer encoder and an 18-layer Transformer decoder, totaling 1 billion parameters.
    \item \textbf{OWSM-CTC v3.1} \cite{peng-etal-2024-owsm}: 27-layer E-Branchformer encoder using CTC for faster inference and reduced hallucination, totaling 1 billion parameters. OWSM-CTC v3.1 uses identical training data to OWSM v3.1.
    \item \textbf{Whisper} \cite{Rad23whisper} In single-corpus experiments, we report results on fine-tuning Whisper models for a fair comparison.
\end{itemize} 

\vspace{-5pt}
\subsection{Front-end representations}
\vspace{-5pt}
To investigate how different front-end representations impact ASR performance, we experiment with both traditional and SSL feature extraction methods:

\begin{itemize}
    \item \textbf{Filterbank (Fbank)}: Mel-scale triangular filtering applied to short-time Fourier transform features (computed using 512-point FFT and 400-sample window).
    \item \textbf{WavLM} \cite{chen2022wavlm}: Representations extracted from WavLM Large (24 Transformer layers), a model pre-trained on 94k hours of primarily English adult speech.
    \item \textbf{XEUS} \cite{chen2024towards}: Representations extracted from XEUS, a model with 19 E-Branchformer layers, pre-trained on 1.1M hours of multilingual speech covering 4000+ languages. 
\end{itemize}
\vspace{-5pt}
\subsection{Continuous and discrete features}
\vspace{-5pt}
Continuous features from pre-trained speech models capture rich acoustic and semantic details, while discrete features obtained via clustering (e.g., K-means), offer compact tokenized representations useful for computationally constrained scenarios. We use a 12 layer E-Branchformer CTC model to compare both approaches on child ASR.
\vspace{-5pt}
\subsection{Single-corpus vs. multi-corpus training}
\vspace{-5pt}
Training on a single corpus allows the model to adapt specifically to the target domain, while multi-corpus training enhances generalizability across domains and helps mitigate data limitations. We train separate models on individual child speech corpora, as listed in Table~\ref{tab:corpora}, as single-corpus models and also train models on a combined dataset that includes all these corpora as multi-corpus models.
\vspace{-5pt}
\subsection{Model scaling}
\vspace{-5pt}
To examine the relationship between model size and performance and determine the optimal balance between complexity and accuracy for child ASR, we conduct zero-shot evaluations of Open Whisper-style
Large-scale neural model Suite (OWLS) ranging from 0.25 billion to 9 billion parameters \cite{chen2025owlsscalinglawsmultilingual}, trained on 180K hours of public speech data. Additionally, we evaluate Whisper models across varying sizes (39M–1.55 billion parameters, tiny to large) to provide a comparative baseline. Since Whisper is trained on 680K hours of speech data, but its exact training details remain undisclosed, its evaluation carries potential risks of bias and data leakage. Whisper large-v3 is excluded due to differences in training data composition compared to smaller Whisper models.
\vspace{-5pt}
\subsection{Age-based ASR performance comparison}
\vspace{-5pt}
We investigate ASR performance across three age groups: 4–7, 8–10, and 11–15. We train models across all ages, and evaluate models separately for each age group to examine how age affects ASR accuracy. Additionally, we also perform zero-shot evaluation of pre-trained models of varying sizes, to analyze trends between model scaling and age-related performance. This setup allows us to determine whether different age groups benefit more from specific model configurations or training paradigms.
\vspace{-5pt}
\subsection{Speaker verification}
\vspace{-5pt}
We also investigate speaker verification performance across age groups. Employing a RawNet3  \cite{jung22_interspeech} within ESPnet framework~\cite{jung24c_interspeech}, we trained the system with AM-Softmax loss in a multi-corpus setting. 

The evaluation focuses on three age groups: 4–7, 8–10, and 11–15 years. 
Performance is measured via Equal Error Rate (EER) and minDCF. These analyses offer insights into child speech processing challenges and the adaptability of speaker embeddings to age variation.

\vspace{-5pt}
\section{Experimental setup}
\vspace{-5pt}
\subsection{Child speech corpora}
\vspace{-5pt}
\begin{table}[!tb]
  \caption{Summary of the child speech corpora used in this study detailing age range, duration, utterance counts, and number of speakers}
  \label{tab:corpora}
  \centering
\scriptsize
\setlength{\tabcolsep}{3pt} 
\renewcommand{\arraystretch}{1.0}
  \begin{tabular}{c c c c c c c}
    \toprule
    \textbf{Corpus} & \shortstack{\textbf{Age Range}\\\textbf{(Years)}} & \shortstack{\textbf{Duration }\\\textbf{(Hours)}}
    & \shortstack{\textbf{Train}\\\textbf{ Utts}} & \shortstack{\textbf{Dev}\\\textbf{Utts}}
    & \shortstack{\textbf{Test}\\\textbf{Utts}} & \textbf{Speakers} \\
    \midrule
    MyST \cite{pradhan-etal-2024-science}    & 8--10   & 179 & 55,702 & 9,037  & 10,311 & 1,371 \\
    OGI Script \cite{shobaki2000ogi}  & 5--15   & 70  & 50,009 & 5,426  & 15,945 & 1,118 \\
    OGI Spon \cite{shobaki2000ogi} & 5--15   & 31  & 3,534  & 349   & 1,095  & 1,101 \\
    CMU Kids \cite{eskenazi1997cmu}       & 6--11   & 9   & 4,468  & 237   & 475   & 76   \\
    \bottomrule
  \end{tabular}
  \vspace{15pt}
\end{table}

The experiments are conducted on three child speech corpora summarized in Table~\ref{tab:corpora}.

The \textbf{MyST} corpus \cite{pradhan-etal-2024-science} comprises dialogues between elementary school students and virtual tutors across 8 science topics. Transcriptions include verbatim orthographic annotations, capturing hesitations, repetitions, and disfluencies. To ensure data quality, we filter out utterances with WER higher than 50\% on Whisper-largeV2, duration outside 0.5-30s,  and fewer than 3 words following \cite{attia2024kid}.

The OGI Kids corpus \cite{shobaki2000ogi} contains scripted and spontaneous speech collected in classroom settings. The scripted portion includes isolated words, sentences, and numeric strings, transcribed as intended target text, without explicit annotation of hesitations or mispronunciations. The spontaneous portion consists of responses to open-ended questions, where transcriptions preserve disfluencies. We split the scripted portion (\textbf{OGI Script}) 70/7.5/22.5\% for train/dev/test, similar to prior works. The spontaneous portion (\textbf{OGI Spon}) is preprocessed using CTC-based forced alignment with Wav2vec2.0 Base \cite{baevski2020wav2vec}, segmented into 25s chunks with 0.5s buffers, and split 71/7/22\%.

The \textbf{CMU Kids} corpus \cite{eskenazi1997cmu} contains read speech from Weekly Reader materials. Transcriptions include the intended target text without explicit annotation of hesitations or mispronunciations. As the official splits contain prompt overlap, we create new splits (7.9/0.4/0.8 hours for train/dev/test) to ensure speaker and prompt independence. 

We note that each dataset follows distinct transcription conventions for mispronunciations and disfluencies making it challenging for multi-corpus models to generalize across annotation styles. Future work may explore normalization strategies to improve ASR hypothesis consistency.
\vspace{-5pt}
\subsection{Implementation details}
\vspace{-5pt}
All experiments are conducted using the ESPnet speech processing framework \cite{watanabe18_interspeech}. To maintain a controlled evaluation, only SpecAugment is applied, without additional augmentation (speed, pitch, etc.) methods. Training utilizes the Adam optimizer with warmup scheduling. To ensure fair comparison, flat-start CTC systems are evaluated with and without external language models, while AED and RNNT use their internal language modeling. All decoding uses greedy search. Throughout the paper, we use Word Error Rate (WER) as the primary metric to measure hypothesis accuracy for ASR models. Most experiments are conducted on GPUs with specifications comparable to V100-32GB, with H100-80GB for OWSM fine-tuning.
\vspace{-5pt}
\section{Results and discussion}
\vspace{-5pt}
\subsection{Flat-start training vs. fine-tuning}
\vspace{-5pt}
Table \ref{tab:flatstart_ft} presents a comparison of flat-start training and fine-tuning, where each model is trained and evaluated on a single dataset. Among flat-start models with Fbank inputs, RNNT achieves the best performance on MyST and OGI Spon, while CTC excels on OGI Script, likely due to the limited vocabulary of the dataset. External language modeling generally improves CTC performance, though benefits vary by dataset. However, most flat-start models lag behind fine-tuned ones, highlighting the challenge of learning child speech from limited data. OWSM v3.1 AED and CTC outperform flat-start models on OGI Script, OGI Spon, and CMU Kids. Whisper-small achieves the lowest WER among 6 of 8 datasets, showcasing the benefits of large-scale (680K hours) pre-training. WER varies across datasets, with OGI Spon being the most challenging and OGI Script yielding the lowest WER, reinforcing the difficulty of spontaneous child speech and the predictability of scripted speech.
\vspace{-5pt}
\subsection{Comparison of front-end representations}
\vspace{-5pt}
We compare E-Branchformer AED with different front-ends (Fbank vs. WavLM and XEUS) under single- and multi-corpus training in Tables \ref{tab:flatstart_ft} and \ref{tab:multicorpus} respectively. For single-corpus training, WavLM and XEUS (rows 2-5, 2-6) show advantages to Fbank (rows 2-1 to 2-3). 
However, in multi-corpus training, the performance gap narrows, and in some cases, WavLM and XEUS result in higher WER than Fbank (rows 4-1 to 4-3 vs. 4-4, 4-5). This is a notable finding, as SSL-based features typically outperform traditional features in adult ASR tasks with similar fine-tuning data sizes \cite{hsu2021hubert, yang21c_interspeech}. Our results suggest that current SSL models are significantly biased toward adult speech, and that flat-start training with combined children's speech is more effective in mitigating these biases. This observation is crucial for advancing child ASR and should be considered in future SSL model development.

\vspace{-5pt}
\begin{table}[!tb]
    \centering
    \scriptsize
    \setlength{\tabcolsep}{3.5pt} 
    \caption{WERs of single-corpus training for flat-start and fine-tuning approaches, evaluated across different child speech datasets.}
    \label{tab:flatstart_ft}
    \renewcommand{\arraystretch}{1.0}
    \begin{threeparttable}
    \begin{tabular}{cc|cc|cc|cc|cc}
        \toprule
        \multirow{2}{*}{\textbf{ID}} &\multirow{2}{*}{\shortstack{\textbf{Model}\\\textbf{Type}}} & \multicolumn{2}{c|}{\textbf{MyST}} & \multicolumn{2}{c|}{\textbf{OGI Script}} & \multicolumn{2}{c|}{\textbf{OGI Spon}} & \multicolumn{2}{c}{\textbf{CMU Kids}} \\
        && \textbf{dev} & \textbf{test} & \textbf{dev} & \textbf{test} & \textbf{dev} & \textbf{test} & \textbf{dev} & \textbf{test} \\
        \midrule
        \multicolumn{10}{c}{\textit{Flat-start E-Branchformer Fbank}} \\
        \midrule
        \rownumber &AED  & 14.5 & 14.8 & 8.8 & 9.7 & 94.6 & 96.1 & 41.4 & 41.6 \\
        \rownumber &RNNT & 12.2 & 12.7 & 8.3 & 8.1 & 58.2 & 59.3 & 43.4 & 43.9 \\
        \rownumber &CTC (no LM) & 16.4 & 16.5 & 2.2 & 2.9 & 73.7 & 73.5 & 41.3 & 42.6 \\
        \rownumber &CTC (with LM) & 16.5 & 16.4 & 1.9 & 2.4 & 74.3 & 73.9 & 40.4 & 41.9 \\
        \midrule
        \multicolumn{10}{c}{\textit{Flat-start E-Branchformer AED}} \\
        \midrule
         \rownumber &WavLM & 9.7 & 10.9 & 3.5 & 3.8 & 45.9 & 44.9 & 39.6 & 37.7 \\
        \rownumber &XEUS &  11.6 & 13.0 & 3.7 & 3.7 & 49.3 & 48.2 & 40.5 & 39.6 \\
        \midrule
        \multicolumn{10}{c}{\textit{OWSM v3.1 (180K hrs) with Fine-tuning}} \\
        \midrule
        \rownumber &AED (1B) & 11.1 & 12.8 & \textbf{1.0}\tnote{*} & \textbf{1.1}\tnote{*} & 12.9 & 14.3 & 13.4 & 13.1 \\
        \rownumber &CTC (1B) & 15.1 & 16.8 & 3.7 & 3.3 & 15.6 & 16.9 & 13.1 & 12.5 \\
        \midrule
        \multicolumn{10}{c}{\textit{Whisper (680K hrs) with Fine-tuning}} \\
        \midrule
        \rownumber &tiny (39 M)& 11.6 & 11.6 & 2.8 & 3.1 & 17.3 & 19.1 & 10.2 & 9.6 \\
        \rownumber &base (74 M)& 9.1 & 10.4 & 2.2 & 2.5 & 13.3 & 14.5 & 6.5 & 7.3 \\
        \rownumber &small (244 M)& \textbf{8.4}\tnote{*} & \textbf{9.3}\tnote{*} & 4.9 & 1.8 & \textbf{10.4}\tnote{*} & \textbf{11.6}\tnote{*} & \textbf{4.7}\tnote{*} & \textbf{5.0}\tnote{*} \\
        \bottomrule
    \end{tabular}
    \begin{tablenotes}
        \item[*] Statistical significance is confirmed with $p < 0.05$
    \end{tablenotes}
    \end{threeparttable}
    \vspace{-2pt}
\end{table}

\vspace{-5pt}
\subsection{Discrete vs. continuous representations}
\vspace{-5pt}
\setcounter{rownumber}{0}
\begin{table}[!tb]
    \vspace*{3mm}
    \centering
    \scriptsize
    \setlength{\tabcolsep}{2.5pt} 
    \caption{WERs of continuous and discrete representations for flat-start training, evaluated across different child speech datasets.}
    \label{tab:discrete_comparison}
    \renewcommand{\arraystretch}{1.0}
    \begin{threeparttable}
    \begin{tabular}{cc|cc|cc|cc|cc}
        \toprule
        \multirow{2}{*}{\textbf{ID}} &\textbf{\multirow{2}{*}{\shortstack{Input\\Features}}} & \multicolumn{2}{c|}{\textbf{MyST}} & \multicolumn{2}{c|}{\textbf{OGI Script}} & \multicolumn{2}{c|}{\textbf{OGI Spon}} & \multicolumn{2}{c}{\textbf{CMU Kids}} \\
        && \textbf{dev} & \textbf{test} & \textbf{dev} & \textbf{test} & \textbf{dev} & \textbf{test} & \textbf{dev} & \textbf{test} \\
        \midrule
            \rownumber & Continuous WavLM & \textbf{9.6}\tnote{*} & \textbf{10.1}\tnote{*} & \textbf{1.6}\tnote{*} & \textbf{1.7}\tnote{*} & \textbf{16.8}\tnote{*} & \textbf{17.2}\tnote{*} & \textbf{26.7}\tnote{*} & \textbf{28.0}\tnote{*} \\
        \rownumber & Discrete WavLM & 10.1 & 10.7 & 2.8 & 3.1 & 27.6 & 27.7 & 44.2 & 44.8 \\
        \bottomrule
    \end{tabular}
    \begin{tablenotes}
        \item[*] Statistical significance is confirmed with $p < 0.05$
    \end{tablenotes}
    \end{threeparttable}
    \vspace{15pt}
\end{table}
Table \ref{tab:discrete_comparison} compares continuous and discrete representations for child speech ASR across multiple datasets. For consistency, we use the last-layer representation of fine-tuned in-domain WavLM as the continuous feature and as input for K-means tokenization (2000 clusters). Continuous WavLM representations yield the lowest WERs across all datasets. Discrete tokens from WavLM perform well on MyST but degrade on other datasets, likely due to limited data and imbalanced sound unit distributions in K-means training, leading to higher information loss. Future work may explore variable clustering, as children’s speech may require different granularity than adult speech.

\vspace{-5pt}
\subsection{Single-corpus vs. multi-corpus training}
\vspace{-5pt}
\setcounter{rownumber}{0}
\begin{table}[!tb]
    \centering
    \scriptsize
    \setlength{\tabcolsep}{3.5pt} 
    \caption{WERs for multi-corpus training for flat-start and fine-tuning approaches, evaluated across different child speech datasets.}
    \label{tab:multicorpus}
    \renewcommand{\arraystretch}{1.0}
    \begin{threeparttable}
    \begin{tabular}{cc|cc|cc|cc|cc}
        \toprule
        \multirow{2}{*}{\textbf{ID}} &\multirow{2}{*}{\shortstack{\textbf{Model}\\\textbf{Type}}} & \multicolumn{2}{c|}{\textbf{MyST}}  & \multicolumn{2}{c|}{\textbf{OGI Script}} & \multicolumn{2}{c|}{\textbf{OGI Spon}} & \multicolumn{2}{c}{\textbf{CMU Kids}} \\
       && \textbf{dev} & \textbf{test} & \textbf{dev} & \textbf{test} & \textbf{dev} & \textbf{test} & \textbf{dev} & \textbf{test} \\
        \midrule
        \multicolumn{10}{c}{\textit{Flat-start E-Branchformer Fbank}} \\
        \midrule
        \rownumber &AED & 12.7 & 13.5 & 2.2 & 2.6 & 24.5 & 24.2 & 15.5 & 16.4  \\
        & \hspace{-10pt} +upsample\tnote{$\dagger$} & 12.5 & 13.1 & \textbf{1.5}\tnote{*} & \textbf{1.9} & 22.2 & 22.0 & 13.3 & 14.0 \\
        \rownumber &RNNT  & 12.2 & 12.6 & 2.3 & 2.7 & 22.4 & 22.1 & 15.7 & 15.1 \\
        \rownumber &CTC (no LM)  & 15.8 & 16.0 & 3.1 & 4.1 & 29.6 & 29.2 & 20.5 & 21.8  \\
        \rownumber &CTC (with LM)  & 15.9 & 15.9 & 3.1 & 4.2 & 29.6 & 28.9 & 19.2 & 20.6  \\
        \midrule
        \multicolumn{10}{c}{\textit{Flat-start E-Branchformer AED}} \\
        \midrule
         \rownumber &WavLM & 11.7 & 14.2 & 7.4 & 10.4 & 21.3 & 20.9 & 10.9 & 11.8  \\
        \rownumber &XEUS & 12.2 & 13.2 & 4.8 & 6.4 & 24.2 & 25.4 & 16.0 & 16.0 \\
        \midrule
        \multicolumn{10}{c}{\textit{OWSM v3.1 Fine-tuning}} \\
        \midrule
        \rownumber &AED & \textbf{9.1}\tnote{*} & \textbf{10.9}\tnote{*} & 1.9 & 2.0 & \textbf{13.7}\tnote{*} & \textbf{14.4}\tnote{*} & \textbf{9.0}\tnote{*} & \textbf{7.5}\tnote{*}  \\
        \rownumber & CTC & 10.5 & 12.5 & 2.3 & 2.8 & 18.1 & 17.8 & 9.9 & 9.0 \\
        \bottomrule
    \end{tabular}
    \begin{tablenotes}
        \item[$\dagger$] Upsample training data from speakers in age 4-7 by 4 and those in age 11-15 by 2, resulting in 61\% more training data.
        \item[*] Statistical significance is confirmed with $p < 0.05$
    \end{tablenotes}
    \end{threeparttable}
\end{table}
Tables \ref{tab:flatstart_ft} (rows 2-1 to 2-8) and \ref{tab:multicorpus} present a comparison of single-corpus and multi-corpus training approaches. Overall, multi-corpus training enhances generalization across most corpora and models. Flat-start Fbank models show notable improvements (rows 2-1 to 2-4 vs. 4-1 to 4-4), while fine-tuned models also benefit from broader training data (rows 2-7, 2-8 vs. 4-7, 4-8). However, for flat-start models using learned representations (rows 2-5, 2-6 vs. 4-5, 4-6), performance trends vary by dataset showing significant gains in some cases but slight degradation in others. To address data imbalance across speakers, we upsampled training data from underrepresented datasets, resulting in improved WER. This highlights the impact of speaker distribution on model performance and suggests that balancing speaker contributions through upsampling can further enhance child ASR robustness. These findings reinforce the need for refined training strategies, as multi-corpus training alone does not fully mitigate dataset composition effects.

\vspace{-5pt}
\subsection{Model scaling}
\begin{figure}[!tb]
  \centering
  \includegraphics[width=0.9\linewidth]{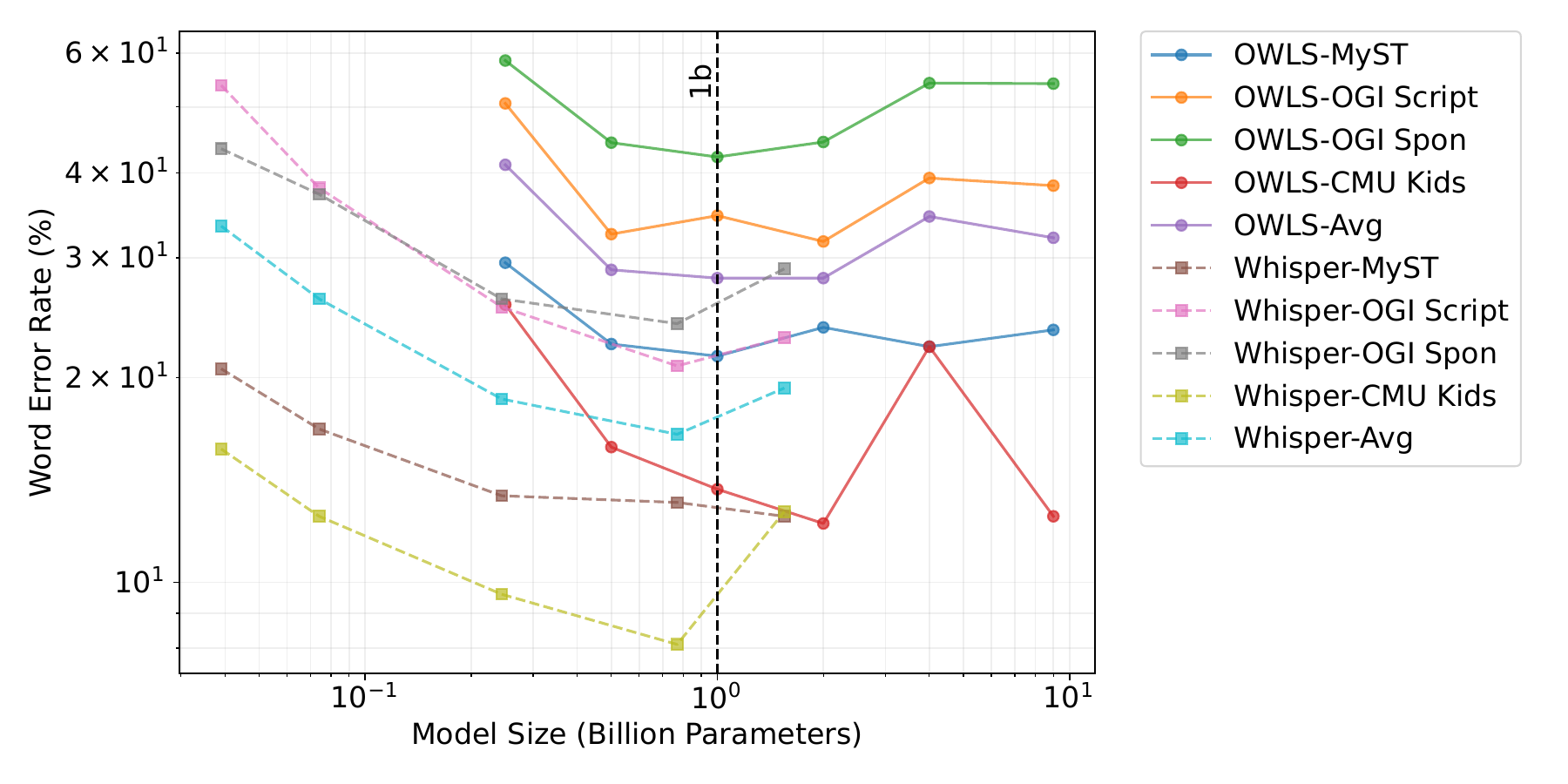}
  \vspace{1pt}
  \caption{Zero-shot WER Scaling of OWLS and Whisper Models}
  \label{fig:scale-wer}
  \vspace{20pt}
\end{figure}

\vspace{-5pt}
Figure \ref{fig:scale-wer} presents a comparison of zero-shot performance across child speech corpora using OWLS (0.25B-9B) and Whisper models (39M-1.55B). The Whisper medium model (769M) achieves the best performance in 6 out of 8 datasets and yields the lowest average WER. We observe an inflection point where WER decreases with model scaling up to $\sim$1B parameters, beyond which performance plateaus or degrades. This trend holds across age groups (4–7, 8–10, 11–15), with both Whisper and OWLS performing optimally at intermediate sizes (0.5B–1.5B). These findings confirm that scaling reliably improves performance up to around 1B parameters, after which larger models may require specialized strategies for child speech.

\vspace{-5pt}
\subsection{ASR performance for different age groups}
\vspace{-5pt}
Table \ref{tab:age_comparison} compares WER across three age ranges: 4–7, 8–10, and 11–15. The data comes from MyST (ages 8–10) and OGI/CMU Kids (covering all three groups). Similar to Section 4.4, we also upsample training data from ages 4-7, and 11-15 for a more balanced data distribution. As expected from the literature \cite{yeung2018difficulties}, younger children’s speech (4–7 years) presents greater challenges, with higher WER and greater performance variance within the same model series. However, an exception arises with zero-shot Whisper models, showing higher WER for ages 11–15 than 8–10. While this discrepancy suggests a strong dependence on Whisper's training data composition, its undisclosed training details make it difficult to pinpoint the exact cause. This observation underscores the limitations of relying on models with closed or proprietary training data when analyzing children's speech. For more reliable insights, models trained on open datasets such as OWSM offer a more interpretable and generalizable approach.

\setcounter{rownumber}{0}
\begin{table}[!tb]
    \centering
    \scriptsize
    \setlength{\tabcolsep}{1.25pt} 
    \caption{WERs for different age groups (4–7, 8–10, 11–15) between flat-start, fine-tuned, and zero-shot pre-trained models.}
    \label{tab:age_comparison}
    \renewcommand{\arraystretch}{1.0}
    \begin{threeparttable}
    \begin{tabular}{ccc|ccc}
        \toprule
        \textbf{ID}&\textbf{Model Type} & \textbf{Approach}  & \textbf{Age 4-7} & \textbf{Age 8-10} & \textbf{Age 11-15} \\
        \midrule
        \rownumber & \multirow{4}{*}{\shortstack{Flat-start Fbank\\ E-Branchformer }} &AED & 10.5 & 10.4 & 2.8 \\
        &    & +upsample\tnote{$\dagger$} & \textbf{7.8}\tnote{*} & 10.0 & 2.3 \\
       \rownumber & &RNNT & 10.5 & 10.2 & 2.8 \\
        \rownumber & &CTC (no LM) & 15.2 & 13.3 & 3.8 \\
        \rownumber & &CTC (with LM) & 12.3 & 15.1 & 3.3 \\
        \midrule
        \rownumber & \multirow{2}{*}{\shortstack{Flat-start AED\\ E-Branchformer}} &WavLM & 32.0 & 11.3 & 6.3 \\
        \rownumber & &XEUS & 23.4 & 11.3 & 5.0 \\
        \midrule
        \rownumber & \multirow{2}{*}{\shortstack{OWSM v3.1 \\Fine-tuning}}&AED & 9.0 & \textbf{7.3}\tnote{*} & \textbf{1.8}\tnote{*} \\
        \rownumber & &CTC  & 12.5 & 8.6 & 2.4 \\
        \midrule
        \rownumber & \multirow{2}{*}{\shortstack{Whisper \\ zero-shot}} 
        &medium (0.769 B)& 39.1 & 13.5 & 15.2 \\
        \rownumber &&large (1.55 B)& 43.9 & 13.4 & 18.6 \\
        \midrule
        \rownumber & \multirow{2}{*}{\shortstack{OWLS \\ zero-shot}}
        &0.5B & 57.2 & 23.6 & 23.2 \\
        \rownumber &&1B & 75.2 & 24.7 & 18.9 \\
        \bottomrule
    \end{tabular}
    \begin{tablenotes}
        \item[$\dagger$] Upsample training data from speakers in age 4-7 by 4 and those in age 11-15 by 2, resulting in 61\% more training data.
        \item[*] Statistical significance is confirmed with $p < 0.05$
    \end{tablenotes}
    \end{threeparttable}
    \vspace{5pt}
\end{table}

\vspace{-5pt}
\subsection{Speaker verification across ages}
\vspace{-5pt}
\setcounter{rownumber}{0}
\begin{table}[!tb]
    \centering
    \scriptsize
    \setlength{\tabcolsep}{5pt} 
    \caption{Comparison of speaker verification performance across different age groups (4–7, 8–10, 11–15).}
    \label{tab:sv_age_comparison}
    \renewcommand{\arraystretch}{1.0}
    \begin{tabular}{cc|cc}
        \toprule
        \textbf{ID}&\textbf{Age Range} & \textbf{EER}  & \textbf{minDCF}\\
        \midrule
        \rownumber & 4--7 & 13.2 & 0.723 \\
        \rownumber & 8--10 & 4.40 & 0.281 \\
        \rownumber & 11--15 & 3.73 & 0.198 \\
        \bottomrule
    \end{tabular}
    \vspace{15pt}
\end{table}
Age-related performance differences extend beyond ASR to speaker verification tasks. Using a RawNet3-based model on OGI Script, we compare Equal Error Rate (EER) and minDCF across three age groups: 4–7, 8–10, and 11–15 (Table \ref{tab:sv_age_comparison}). Consistent with prior works \cite{singh2024childaugment}, younger children (4–7 years) show higher EER and minDCF due to greater speaker variability, indicating reduced embedding robustness. These results parallel our ASR findings and emphasize the need for speaker embedding models that better generalize to child speech variability.

\vspace{-5pt}
\section{Conclusion}
Our study comprehensively analyzes child speech processing, examining training strategies, model architectures, front-end representations, and scaling effects. While fine-tuned models generally perform best, flat-start models help mitigate biases in SSL representations, which are predominantly trained on adult speech. Multi-corpus training improves generalization, though dataset composition significantly impacts robustness. Model scaling benefits child ASR up to ~1B parameters before plateauing. Our age-based analysis highlights inconsistencies in Whisper's performance and reveals greater speaker variability among younger children, reinforcing the need for open-data models for reliable evaluation. Future work should refine SSL models for child speech and optimize multi-corpus training strategies. All experiments use a unified ESPnet-based framework, ensuring transparency and supporting future child speech research.

\vspace{-5pt}
\section{Acknowledgement}
Experiments of this work used the Bridges2 system at PSC and Delta system at NCSA through allocations CIS210014 and IRI120008P from the Advanced Cyberinfrastructure Coordination Ecosystem: Services \& Support (ACCESS) program, supported by National Science Foundation grants \#2138259, \#tel:2138286, \#tel:2138307, \#tel:2137603, \#tel:2138296 and \#2202585.
\\ Additional support was provided by FWO-SBO grant S004923N: NEFL, KU Leuven C24M/22/025, and FWO grant V401325N.

\let\oldthebibliography\thebibliography
\def\thebibliography#1{\oldthebibliography{#1}%
\setlength{\itemsep}{-1.2pt}%
\setlength{\parsep}{0pt}%
\setlength{\topsep}{0pt}}

\bibliographystyle{IEEEtran}
\bibliography{mybib}

\end{document}